\documentclass[final]{cvpr}

\usepackage{times}
\usepackage{epsfig}
\usepackage{graphicx}
\usepackage{amsmath}
\usepackage{amssymb}
\usepackage{booktabs}
\graphicspath{{cambriafig/}}
\usepackage[pagebackref=true,breaklinks=true,colorlinks,bookmarks=false]{hyperref}
\usepackage{float}

\newcommand{\bs}{\boldsymbol}
\def\cvprPaperID{2662} % *** Enter the CVPR Paper ID here
\def\confYear{CVPR 2021}
\begin{document}

%%%%%%%%% TITLE
%Recurrent Neural Transporter (RNT)
\title{PMP-Net: Point Cloud Completion by Learning Multi-step Point Moving Paths}

\author{Xin Wen\textsuperscript{1}, Peng Xiang\textsuperscript{1}, Zhizhong Han\textsuperscript{2}, Yan-Pei Cao\textsuperscript{3}, Pengfei Wan\textsuperscript{3}, Wen Zheng\textsuperscript{3}, Yu-Shen Liu\textsuperscript{1}\thanks{Corresponding author. This work was supported by National Key R\&D Program of China (2020YFF0304100, 2018YFB0505400), the National Natural Science Foundation of China (62072268), and in part by Tsinghua-Kuaishou Institute of Future Media Data.}\\
\textsuperscript{1}School of Software, BNRist, Tsinghua University, Beijing, China\\
\textsuperscript{2}Department of Computer Science, Wayne State University, USA\\
\textsuperscript{3} Y-tech, Kuaishou Technology, Beijing, China\\
{\small\{x-wen16,xp20\}@mails.tsinghua.edu.cn\hspace{1mm}
h312h@wayne.edu\hspace{1mm}caoyanpei@gmail.com\hspace{1mm}}\\
{\small\{wanpengfei,zhengwen\}@kuaishou.com\hspace{1mm}
liuyushen@tsinghua.edu.cn}
}

\maketitle
\thispagestyle{empty}

%%%%%%%%% ABSTRACT
\begin{abstract}
The task of point cloud completion aims to predict the missing part for an incomplete 3D shape. A widely used strategy is to generate a complete point cloud from the incomplete one. However, the unordered nature of point clouds will degrade the generation of high-quality 3D shapes, as the detailed topology and structure of discrete points are hard to be captured by the generative process only using a latent code. In this paper, we address the above problem by reconsidering the completion task from a new perspective, where we formulate the prediction as a point cloud deformation process. Specifically, we design a novel neural network, named PMP-Net, to mimic the behavior of an earth mover. It moves each point of the incomplete input to complete the point cloud, where the total distance of point moving paths (PMP) should be shortest.
Therefore, PMP-Net predicts a unique point moving path for each point according to the constraint of total point moving distances.
As a result, the network learns a strict and unique correspondence on point-level,
%which can capture the detailed topology and structure relationships between the incomplete shape and the complete target,
and thus improves the quality of the predicted complete shape.
%and produce the ordered output following the exact arrangement of input point cloud.
We conduct comprehensive experiments on Completion3D and PCN datasets, which demonstrate our advantages over the state-of-the-art point cloud completion methods. Code will be available at \url{https://github.com/diviswen/PMP-Net}.

\end{abstract}

%%%%%%%%% BODY TEXT
\section{Introduction}
As one of the widely used 3D shape representations, point cloud can be easily obtained through depth cameras or other 3D scanning devices. Due to the limitations of view-angles or occlusions of 3D scanning devices, the raw point clouds are usually sparse and incomplete \cite{wen2020sa}. Therefore, a shape completion/consolidation process is usually required to generate the missing regions of 3D shape for the downstream 3D computer vision applications like classification \cite{han20193dviewgraph,han2019parts,wen2020point2spatialcapsule,han2019seqviews2seqlabels,liu2019sequence,liu2020lrc}, segmentation \cite{Wen2020MM,jiang2020pointgroup} and other visual analysis \cite{yuan2021survey}.

In this paper, we focus on the completion task for 3D objects represented by point clouds, where the missing parts are caused by self-occlusion due to the view angle of scanner. Most of the previous methods formulate the point cloud completion as a point cloud generation problem \cite{dai2017shape,wen2020sa,yuan2018pcn,tchapmi2019topnet}, where an encoder-decoder framework is usually adopted to extract a latent code from the input incomplete point cloud, and decode the extracted latent code into a complete point cloud. Benefiting from the deep learning based point cloud learning methods, the point cloud completion methods along this line have made huge progress in the last few years \cite{wen2020sa,tchapmi2019topnet}. However, the generation of point clouds remains a difficult task using deep neural network, because the unordered nature of point clouds makes the generative model difficult to capture the detailed topology or structure among discrete points \cite{tchapmi2019topnet}. Therefore, the completion quality of point clouds based on generative models is still unsatisfying.

\begin{figure}[!t]
  \centering
  \includegraphics[width=\columnwidth]{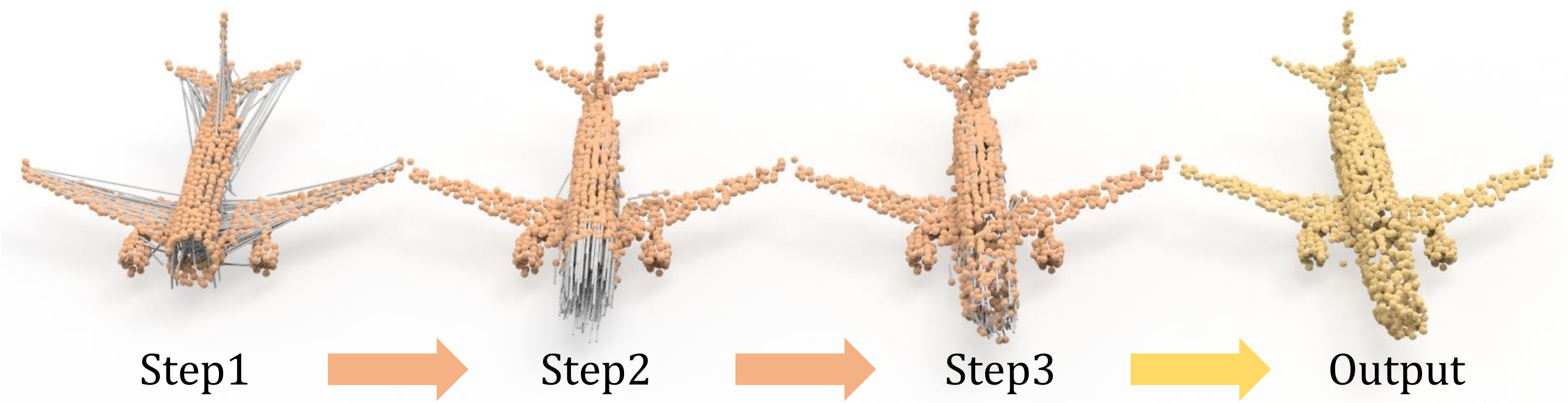}\vspace{-0.4cm}
  \caption{Illustration of completion by multi-step point cloud deformation. The point moving paths are denoted by grey lines. At each step, the source point cloud is colored by orange and the target point cloud is colored by yellow, respectively.
  }
  \label{fig:searching_illustration}
\end{figure}

To address this issue, in this paper, we propose a novel neural network, named PMP-Net, to formulate the task of point cloud completion from a new perspective. Different from the generative model that directly predicts the coordinations of all points in 3D space, the PMP-Net learns to move the points from the source 3D shape to the target one.
%Specifically, taking input incomplete point cloud as the source, PMP-Net works like a path planner to arrange a point moving path for each points in the source point cloud, and move them to the target point cloud.
Through the point moving process, the PMP-Net establishes the point-level correspondences between the source point cloud and the target, which captures the detailed topology and structure relationships between the two point clouds. On the other hand, there are various solutions to move points from source to target, which will confuse the network during training.
Therefore, in order to encourage the network to learn a unique arrangement of point moving path, we take the inspiration from the Earth Mover's Distance (EMD) and  propose to regularize a Point-Moving-Path Network (PMP-Net) under the constraint of the total point moving distances (PMDs), which guarantees the uniqueness of path arrangement between the source point cloud and the target one.
%and eliminate the problem of predicting unordered data (since the EMD constraints establishes a strict and unique correspondence between the input and the output point cloud).

\begin{figure}[!t]
  \centering
  \includegraphics[width=\columnwidth]{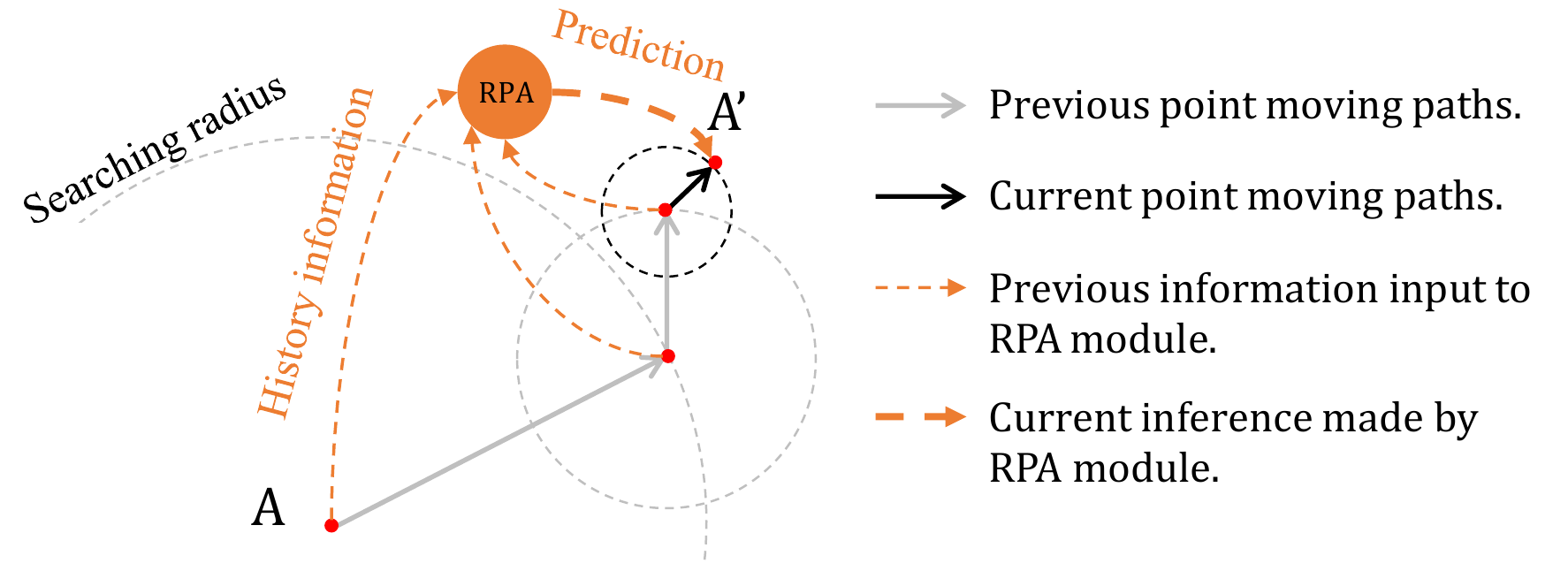}\vspace{-0.1cm}
  \caption{Illustration of path searching with multiple steps under the coarse-to-fine searching radius. The PMP-Net moves point A to point A' by three steps, with each step reducing its searching radius, and looking back to consider the moving history in order to decide the next place to move.
  }
  \label{fig:searching_illustration}
\end{figure}

Moreover, in order to predict the point moving path more accurately, we propose a multi-step path searching strategy to continuously refine the point moving path under multi-scaled searching radius. Specifically, as shown in Figure \ref{fig:searching_illustration}, the path searching is repeated for multiple steps in a coarse-to-fine manner. Each step will take the previously predicted paths into consideration, and then plan its next move path according to the previous paths. To record and aggregate the history information of point moving path, we take the inspiration from Gated Recurrent Unit (GRU) to propose a novel Recurrent Path Aggregation (RPA) module. It can memorize and aggregate the route sequence for each point, and combine the previous information with the current location of point to predict the direction and the length for the next move. By reducing the searching radius step-by-step, PMP-Net can consistently refine a more and more accurate path for each point to move from its original position on the incomplete point cloud to the target position on the complete point cloud.
In all, the main contribution of our work can be summarized as follows.
\begin{itemize}
  \item We propose a novel network for point cloud completion task, named PMP-Net, to move each point from the incomplete shape to the complete one to achieve a high quality point cloud completion. Compared with previous generative completion methods, PMP-Net has the ability to learn more detailed topology and structure relationships between incomplete shapes and complete ones, by learning the point-level correspondence through point moving path prediction.
  \item We propose to learn a unique point moving path arrangement between input and output point clouds, by regularizing the network using the constraint of Earth Mover's Distance. As a result, the network will not be confused by multiple solutions of moving points, and finally predicts a meaningful point-wise correspondence between the source and target point clouds.
  \item We propose to search point moving path with multiple steps in a coarse-to-fine manner. Each step will decide the next move based on the aggregated information from the previous paths and its current location, by using the proposed Recurrent Path Aggregation (RPA) module.
\end{itemize}

\section{Related Work}
The deep learning technology in 3D reconstruction \cite{Han2020TIP,Han2020Sdrwr,Han2020ECCV,Jiang2019SDFDiffDRcvpr,Han2020Sdrwr} and representation learning \cite{9318534,han2019multi,liu2019l2g,han20193d2seqviews} have boosted the research of 3D shape completion, which can be roughly divided into two categories. (1) Traditional 3D shape completion methods \cite{sung2015data,berger2014state,thanh2016field,wei2019local} usually formulate hand-crafted features such as surface smoothness or symmetry axes to infer the missing regions, while some other methods \cite{shao2012interactive,kalogerakis2012probabilistic,martinovic2013bayesian,shen2012structure} consider the aid of large-scale complete 3D shape datasets, and perform searching to find the similar patches to fill the incomplete regions of 3D shapes. (2) Deep learning based methods \cite{zhang2020detail,sarmad2019rl,huang2020pf,hutaoaaai2020}, on the other hand, exploit the powerful representation learning ability to extract geometric features from the incomplete input shapes, and directly infer the complete shape according to the extracted features. Those learnable methods do not require the predefined hand-crafted features in contrast with traditional completion methods, and can better utilize the abundant shape information lying in the large-scale completion datasets. The proposed PMP-Net also belongs to the deep learning based method, where the methods along this line can be further categorized and detailed as below.

\noindent\textbf{Volumetric aided shape completion.}
The representation learning ability of convolutional neural network (CNN) has been widely used in 2D computer vision research, and the studies concerning application of 2D image inpainting have been continuously surging in recent years. A intuitive idea for 3D shape completion can be directly borrowed from the success of 2D CNN in image inpainting research \cite{yu2018generative,yeh2017semantic,liu2018image}, extending it into 3D space. Recently, several volumetric aided shape completion methods, which are based on 3D CNN structure, have been developed. Note that we use the term \emph{``volumetric aided''} to describe this kind of methods, because the 3D voxel is usually not the final output of the network. Instead, the predicted voxel will be further refined and converted into other representations like mesh \cite{dai2017shape} or point cloud \cite{xie2020grnet}, in order to produce more detailed 3D shapes. Therefore, the voxel is more like an intermediate aid to help the completion network infer the complete shape. Notable works along this line like 3D-EPN \cite{dai2017shape} and GRNet \cite{xie2020grnet} have been proposed to reconstruct the complete 3D voxel in a coarse-to-fine manner. They first predict a coarse complete shape using 3D CNN under an encoder-decoder framework, and then refine the output using similar patches selected from a complete shape dataset \cite{dai2017shape} or by further reconstructing the detailed point cloud according to the output voxel \cite{xie2020grnet}. Also, there are some studies that consider purely volumetric data for shape completion task. For example, Han et. al \cite{han2017high} proposed to directly generate the high-resolution 3D volumetric shape, by simultaneously inferring global structure and local geometries to predict the detailed complete shape. Stutz et. al \cite{stutz2018learning} proposed a variational auto-encoder based method to complete the 3D voxel under weak supervision. Despite the fascinating ability of 3D CNN for feature learning, the computational cost which is cubic to the resolution of input voxel data makes it difficult to process fine-grained shapes \cite{wen2020sa}.

\noindent\textbf{Point cloud based shape completion.}
There is a growing attention on the task of point cloud based shape completion \cite{xin2021c4c,tchapmi2019topnet,wen2020sa,hu2019render4completion} in recent years. Since point cloud is a direct output form of many 3D scanning devices, and the storage and process of point clouds require much less computational cost than volumetric data, many recent studies consider to perform direct completion on 3D point clouds. Enlighten from the improvement of point cloud representation learning \cite{qi2017pointnet++,qi2017pointnet}, previous methods like TopNet \cite{tchapmi2019topnet}, PCN \cite{yuan2018pcn} and SA-Net \cite{wen2020sa} formulate the solution as a generative model under an encoder-decoder framework. They adopted encoder like PointNet \cite{qi2017pointnet} or PointNet++ \cite{qi2017pointnet++} to extract the global feature from the incomplete point cloud, and use a decoder to infer the complete point cloud according to the extracted features. Compare to PCN \cite{yuan2018pcn}, TopNet \cite{tchapmi2019topnet} improved the structure of decoder in order to implicitly model and generate point cloud in a rooted tree architecture \cite{tchapmi2019topnet}. SA-Net \cite{wen2020sa} took one step further to preserve and convey the detailed geometric information of incomplete shape into the generation of complete shape through skip-attention mechanism. Other notable work like RL-GAN-Net \cite{sarmad2019rl} and Render4Completion \cite{hu2019render4completion} focused on the framework of adversarial learning to improve the reality and consistency of the generated complete shape. In all, most of the above methods are generative solution for point cloud completion task, and inevitably suffer from the unordered nature of point clouds, which makes it difficult to reconstruct the detailed typology or structure using a generative decoder. Therefore, in order to avoid the problem of predicting unordered data, PMP-Net uses a different way to reconstruct the complete point cloud, which learns to move all points from the initial input instead of directly generating the final point cloud from a latent code.
The idea of PMP-Net is also related to the research of 3D shape deformation \cite{yin2018p2p}, which mainly considered one-step deformation. However, the deformation between the incomplete and complete shapes is more challenging, which requires the inference of totally unknown geometries in missing regions without any other prior information. In contrast, we propose multi-step searching to encourage PMP-Net to infer more detailed geometric information for missing region, along with point moving distance regularization to guarantee the efficiency of multi-step inference.

\begin{figure*}[!t]
  \centering
  \includegraphics[width=\textwidth]{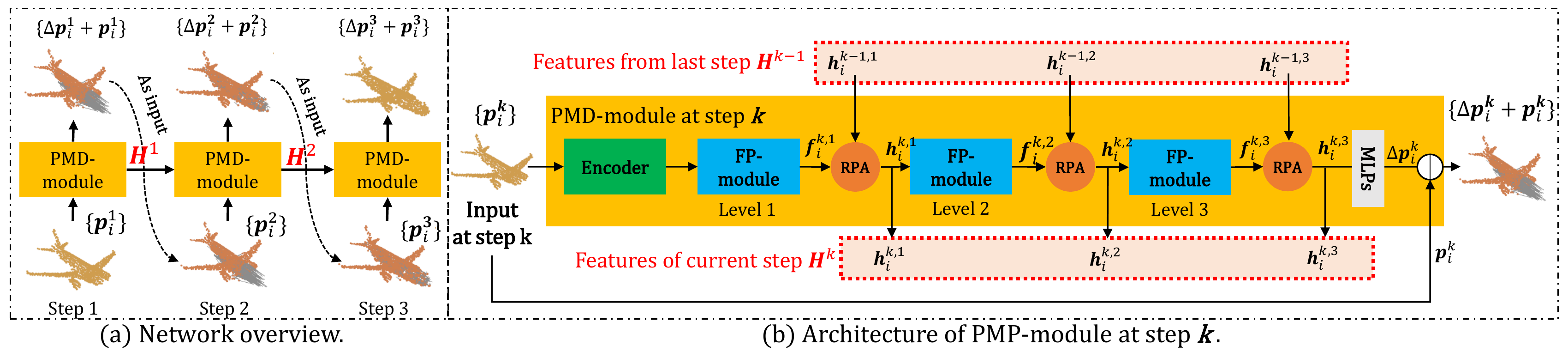}\vspace{-0.4cm}
  \caption{Illustration of path searching with multiple steps under the coarse-to-fine searching radius. The PMP-Net moves point A to point A' by three steps, with each step reducing its searching radius, and looking back to consider the moving history in order to decide the next place to move.
  }
  \label{fig:overview}
\end{figure*}

\section{Architecture of PMP-Net}

%\noindent\textbf{Basic Formulations and Overview}
%Given an input point cloud $P=\{\bs{p}_i\}$ and a target point cloud $P'=\{\bs{p'}_j\}$. The objective of PMP-Net is to predict a displacement vector set $\Delta P=\{\Delta \bs{p}_i\}$, which can move each point from $P$ into the position of $P'$ such that $\{(\bs{p}_i+\Delta \bs{p}_i)\}=\{\bs{p'}_j\}$.

%An overview of the proposed PMP-Net is shown in Figure \ref{fig:overview}(a). The network basically consists of three parts: (1) the encoder to extract point cloud features; (2) the feature propagation module (FP-Module) to predict the point moving path for each point; (c) the RPA module recurrently fuses and integrates the current step's point features with the previous steps' path information. The details of each part will be described as below.

\subsection{Point Displacement Prediction}
\noindent\textbf{Multi-step framework.}
An overview of the proposed PMP-Net is shown in Figure \ref{fig:overview}(a). Given an input point cloud $P=\{\bs{p}_i\}$ and a target point cloud $P'=\{\bs{p'}_j\}$. The objective of PMP-Net is to predict a displacement vector set $\Delta P=\{\Delta \bs{p}_i\}$, which can move each point from $P$ into the position of $P'$ such that $\{(\bs{p}_i+\Delta \bs{p}_i)\}=\{\bs{p'}_j\}$.
PMP-Net moves each point $\bs{p}_i$ for $K=3$ steps in total. The displacement vector for step $k$ is denoted by $\Delta\bs{p}^k_i$, so $\Delta \bs{p}_i=\sum_{k=1}^{3}\Delta\bs{p}^k_i$. For step $k$, the network takes the deformed point cloud $\{\bs{p}^{k-1}_i\}=\{\bs{p}_i+\sum_{j=1}^{k-1}\Delta\bs{p}^j_i\}$ from the last step $k-1$ as input, and calculates the new displacement vector according to the input point cloud. Therefore, the predicted shape will be consistently refined step-by-step, which finally produces a complete shape with high quality.%, as denoted by the black arrow in Figure \ref{fig:overview}(a).

\noindent\textbf{Displacement vector prediction.}
At step $k$, in order to predict the displacement vector $\Delta\bs{p}^k_i$ for each point, we first extract per-point features from the point cloud. This is achieved by first adopting the basic framework of PointNet++ \cite{qi2017pointnet++} to extract the global feature of input 3D shape, and then using the feature propagation module to propagate the global feature to each point in 3D space, and finally producing per-point feature $\bs{h}^{k,l}_i$ for point $\bs{p}^k_i$. Since our experimental implementation applies three levels of feature propagation to hierarchically produce per-point feature (see Figure \ref{fig:overview}(b)), we use superscript $k$ to denote the step and the subscript $l$ to denote the level in $\bs{h}^{k,l}_i$.
The per-point feature $\bs{h}^{k,l}_i$ is then concatenated with a random noise vector $\hat{\bs{x}}$, which according to \cite{yin2018p2p} can give point tiny disturbances and force it to leave its original place. Then, the final point feature $\bs{h}^{k,3}_i$ at step k and level 3 is fed into a multi-layer perceptron (MLP) followed by a hyper-tangent activation (tanh), to produce a 3-dimensional vector as the displacement vector $\Delta\bs{p}^k_i$ for point $\bs{p}^k_i$ as
\begin{equation}\label{eq:displacement}\small
  \Delta\bs{p}^k_i = \mathrm{tanh}(\mathrm{MLP}([\bs{h}^{k,3}_i:\hat{\bs{x}}])),
\end{equation}
where ``:'' denotes the concatenation operation.

\noindent\textbf{Recurrent information flow between steps.}
The information of previous moves is crucial for network to decide the current move, because the previous paths can be used to infer the location of the final destination for a single point. Moreover, such information can guide the network to find the direction and distance of next move, and prevent it from changing destination during multiple steps of point moving path searching. In order to achieve this target, we propose to use a special RPA unit between each step and each level of feature propagation module, which is used to memorize the information of previous path and to infer the next position of each point. As shown in Figure \ref{fig:overview}(b), the RPA module in (step $k$, level $l$) takes the output $\bs{f}^{k,l-1}_i$ from the last level i-1 as input, and combines it with the feature $\bs{h}^{k-1,l}_i$ from the previous step $k-1$ at the same level $l$ to produce the feature of current level $\bs{h}^{k,l}_i$, denoted as
\begin{equation}\small
  \bs{h}^{k,l}_i = \mathrm{RPA}(\bs{f}^{k,l-1}_i, \bs{h}^{k-1,l}_i).
\end{equation}
The detailed structure of RPA module is described below.

\begin{figure}[!t]
  \centering
  \includegraphics[width=\columnwidth]{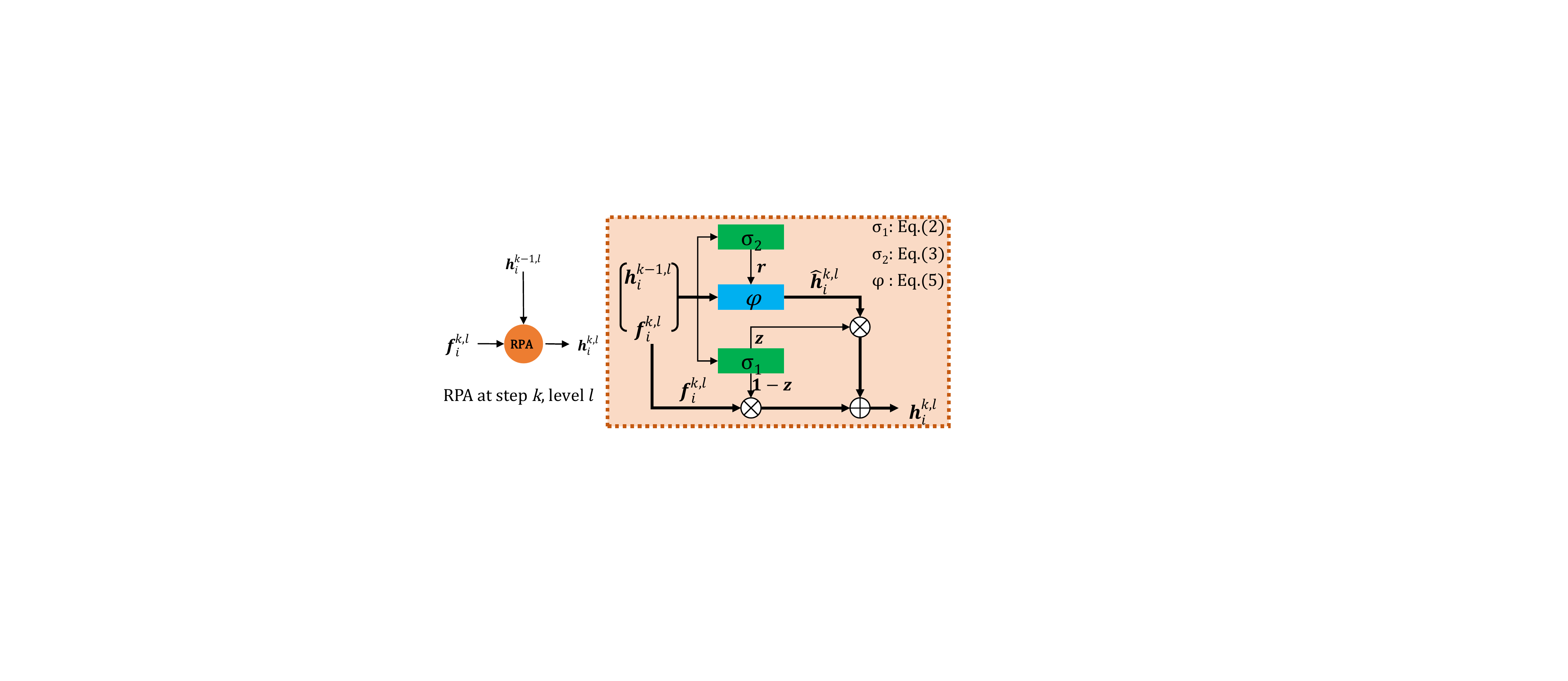}\vspace{-0.4cm}
  \caption{Detailed structure of RPA module at step $k$, level $l$.
  }
  \label{fig:rpa_structure}
\end{figure}

\subsection{Recurrent Path Aggregation}
The previous paths of point moving can be regarded as the sequential data, where the information of each move should be selectively memorized or forgotten during the process.
%Therefore, the RPA unit should be capable of encoding current information, and also be able to selectively forget early information.
Following this idea, we take the inspiration from the recurrent neural network, where we mimic the behavior of gated recurrent unit (GRU) to calculate an update gate $\bs{z}$ and reset gate $\bs{r}$ to encode and forget information, which is according to the point feature $\bs{h}^{k-1,l}_{i}$ from the last step $k-1$ and the point feature $\bs{f}^{k,l-1}_{i}$ of current step $k$. The calculation of two gates can be formulated as
\begin{equation}\small
  \bs{z} = \sigma(W_z[\bs{f}^{k,l-1}_{i}:\bs{h}^{k-1,l}_{i}]+\bs{b}_z),
\end{equation}
\begin{equation}\small
  \bs{r} = \sigma(W_r[\bs{f}^{k,l-1}_{i}:\bs{h}^{k-1,l}_{i}]+\bs{b}_r),
\end{equation}
where ${W_z, W_r}$ are weight matrix and ${\bs{b}_z,\bs{b}_z}$ are biases. $\sigma$ is the \emph{sigmoid} activation function, which predicts a value between 0 and 1 to indicate the ratio of information that allowed to pass the gate. ``:'' denotes the concatenation of two features.

Different from the standard GRU, which emphasizes more importance on the preservation of previous information when calculating the output feature $\bs{h}^{k,l}_{i}$ at current step, in RPA, we address more importance on the preservation of current input information, and propose to calculate the output feature $\bs{h}^{k,l}_{i}$ as
\begin{equation}\label{eq:hi}\small
  \bs{h}^{k,l}_{i} = \bs{z}\odot\bs{\hat{h}}^{k,l}_{i}+(1-\bs{z})\odot \bs{f}^{k,l-1}_{i},
\end{equation}
where $\bs{\hat{h}}^{k,l}_{i}$ is the intermediate feature of current step. It contains the preserved information from the past, which is calculated according to the current input feature. The formulation of $\bs{\hat{h}}^{k,l}_{i}$ is given as
\begin{equation}\small
  \bs{\hat{h}}^{k,l}_{i} = \varphi(W_h[\bs{r}\odot\bs{h}^{k-1,l}_{i}: \bs{f}^{k,l-1}_{i}]+\bs{b}_h),
\end{equation}
where $\varphi$ is \emph{relu} activation in our implementation.

The reason of fusing $\bs{\hat{h}}^{k,l}_{i}$ with $\bs{f}^{k,l-1}_{i}$ instead of $\bs{h}^{k-1,l}_{i}$ is that, compared with standard unit in RNN unit, the current location of point should have greater influence to the decision of next move. Especially, when RPA module needs to ignore the previous information which is not important in the current decision making, Eq.(\ref{eq:hi}) can easily allow RPA model to forget all history by simply pressing the update gate $\bs{z}$ to a zero-vector, and thus enables the RPA module fully focus on the information of current input $\bs{f}^{k,l-1}_{i}$.

\subsection{Optimized Searching for Unique Paths}

\begin{figure}[!t]
  \centering
  \includegraphics[width=\columnwidth]{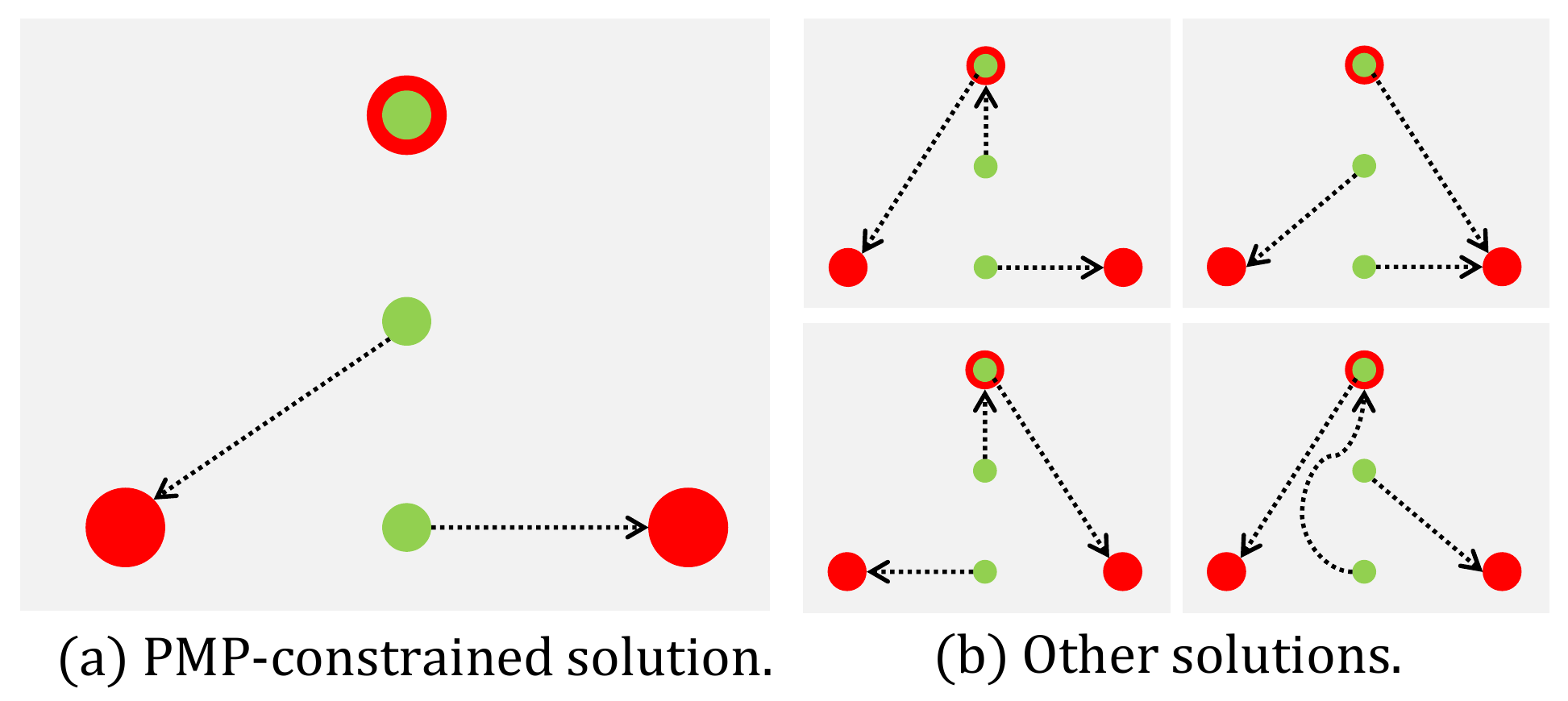}\vspace{-0.4cm}
  \caption{Illustration of multiple solutions when deforming input point cloud (green) into target point cloud (red). The PMD-constraint guarantees the uniqueness of point level correspondence (a) between input and target point cloud, and filter out various redundant solutions for moving points (b).
  }
  \label{fig:emd_solution}
\end{figure}

\noindent\textbf{Minimizing moving distance.}
As shown in Figure \ref{fig:emd_solution}, the unordered nature of point cloud allows multiple solutions to deform the input shape into the target one, and the direct constraint (e.g. chamfer distance) on the deformed shape and its ground truth cannot guarantee the uniqueness of correspondence established between the input point set and the target point set. Otherwise, the network will be confused by the multiple solutions of point moving, which may lead to the failure of capturing detailed topology and structure relationships between incomplete shapes and complete ones.
In order to establish a unique and meaningful point-wise correspondence between input point cloud and target point cloud, we take the inspiration from Earth Mover's Distance \cite{rubner2000earth}, and propose to train PMP-Net to learn the path arrangement $\phi$ between source and target point clouds under the constraint of total point moving path distance. Specifically, given the source point clouds $\hat{X}=\{\hat{\bs{x}}_i|i=1,2,3,...,N\}$ and the target point cloud $X=\{\bs{x}_i|i=1,2,3,...,N\}$, we follow EMD to learn an arrangement $\phi$ which meets the constraint below
\begin{equation}\label{eq:emd}\small
  \mathcal{L}_{\mathrm{EMD}}(\hat{X},X)=\min_{\phi :\hat{X}\rightarrow X}\frac{1}{\hat{X}}\sum_{\hat{\bs{x}}\in\hat{X}}\| \hat{\bs{x}}-\phi(\hat{\bs{x}}) \|,
\end{equation}
In Eq.(\ref{eq:emd}), $\phi$ is considered as a bijection that minimizes the average distance between corresponding points in $\hat{X}$ and $X$.

\begin{figure}[!t]
  \centering
  \includegraphics[width=\columnwidth]{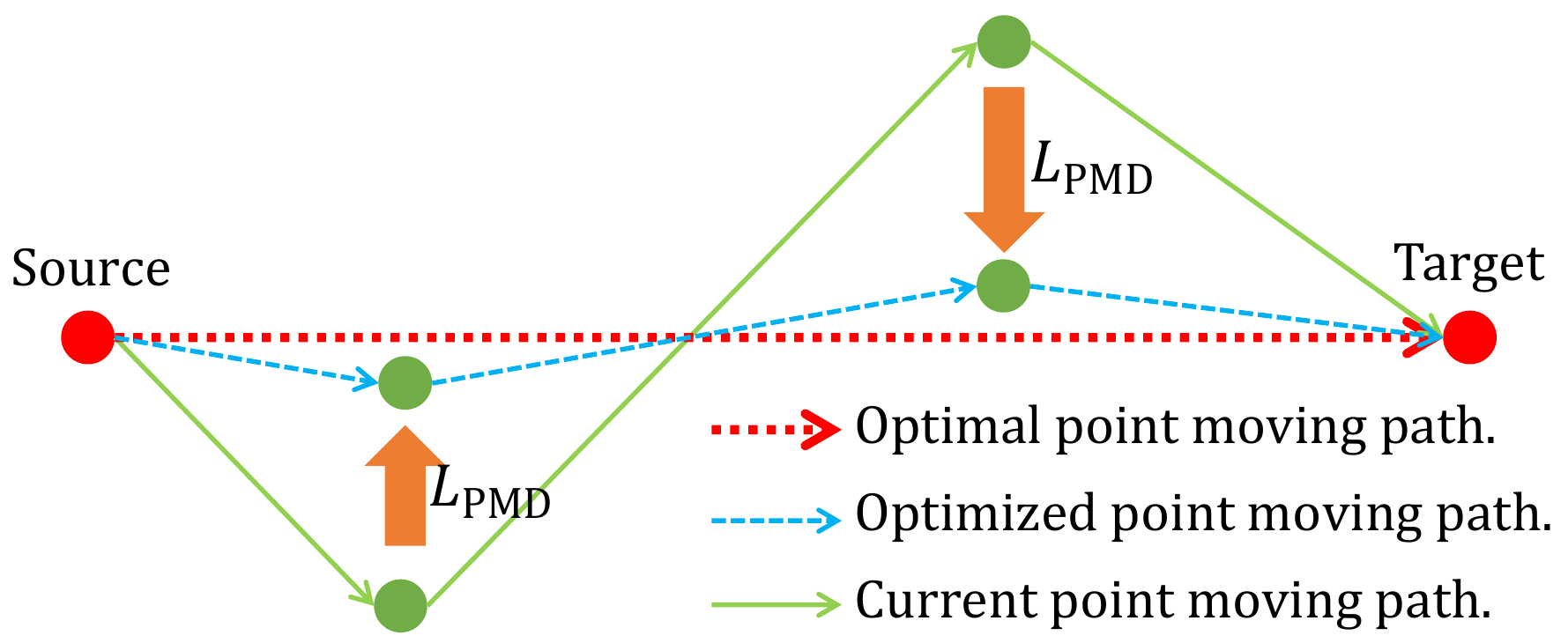}\vspace{-0.4cm}
  \caption{Illustration of the effectiveness of $\mathcal{L}_{PMD}$. By minimizing the point moving distance, the network is encouraged to learn more consistent paths from source to target, which will reduce redundant searching in each step and improve the efficiency.
  }
  \label{fig:mdl}
\end{figure}

According to Eq.(\ref{eq:emd}), bijection $\phi$ established by the network should achieve the minimum moving distance to move points from input shape to target shape. However, even if the correspondence between input and target point clouds is unique, there still exist various paths between source and target points, as shown in Figure \ref{fig:mdl}. Therefore, in order to encourage the network to learn an optimal point moving path, we choose to minimize the point moving distance loss ($\mathcal{L}_{\mathrm{PMD}}$), which is the sum of all displacement vector $\{\Delta \bs{p}_i^k\}$ output by all three steps in PMP-Net. %This encourages the network to search a path that satisfy the requirements of Eq.\ref{eq:emd}, and output a unique path arrangements with meaningful correspondence.
The \emph{Point Moving Distance} loss is formulated as
\begin{equation}\label{eq:mdl}\small
  \mathcal{L}_{\mathrm{PMD}} = \sum_{k}\sum_{i}\|\Delta \bs{p}^k_i \|_2.
\end{equation}

Eq.(\ref{eq:mdl}) is more strict than EMD constraint. It requires not only the overall displacements of all point achieve the shortest distance, but also limits the point moving paths in each step to be the shortest one. Therefore, in each step, the network will be encouraged to search new path following the previous direction, as shown in Figure \ref{fig:mdl}, which will lead to less redundant moving decision and improve the searching efficiency.

\begin{table*}[!t]\small
\centering
\caption{Point cloud completion on Completion3D dataset in terms of per-point L2 Chamfer distance $\times 10^{4}$ (lower is better).}
%\resizebox{\linewidth}{!}{
\begin{tabular}{l|c|cccccccc}
\toprule
Methods &Average  &Plane    &Cabinet  &Car   &Chair   &Lamp   &Couch    &Table    &Watercraft      \\ \midrule
FoldingNet  \cite{yang2018foldingnet}   &19.07  &12.83    &23.01    &14.88    &25.69    &21.79    &21.31    &20.71    &11.51   \\
PCN  \cite{yuan2018pcn}   &18.22 &9.79    &22.70    &12.43    &25.14    &22.72    &20.26    &20.27    &11.73   \\
PointSetVoting \cite{zhang2020point}  &18.18 &6.88    &21.18    &15.78    &22.54    &18.78    &28.39    &19.96    &11.16   \\
AtlasNet \cite{groueix2018atlasnet}   &17.77   &10.36    &23.40    &13.40    &24.16    &20.24    &20.82    &17.52    &11.62   \\
SoftPoolNet \cite{wang2020softpoolnet} &16.15   &5.81   &24.53   &11.35    &23.63    &18.54      &20.34   &16.89  &7.14   \\
TopNet  \cite{tchapmi2019topnet}     &14.25   &7.32   &18.77   &12.88    &19.82    &14.60      &16.29   &14.89  &8.82   \\
SA-Net \cite{wen2020sa}  &11.22   &5.27   &\textbf{14.45}   &\textbf{7.78}    &13.67    &13.53      &14.22   &11.75  &8.84   \\
%GRNet(2k) \cite{xie2020grnet}  &11.22   &5.7   &16.51   &8.63    &14.48    &11.36      &15.50   &10.67  &6.19   \\
GRNet \cite{xie2020grnet}  &10.64   &6.13   &16.90   &8.27    &12.23    &10.22      &14.93   &10.08  &5.86   \\
\midrule
PMP-Net(Ours) &\textbf{9.23}  &\textbf{3.99}    &14.70    &8.55    &\textbf{10.21}    &\textbf{9.27}    &\textbf{12.43}    &\textbf{8.51}   &\textbf{5.77} \\
\bottomrule
\end{tabular}
%}
\label{table:completion3D}
\end{table*}

%Here, we give a concise discussion about the effect of moving distance loss: $\mathcal{L}_{mdl}$ is a relatively loose constraints which requires the network to minimize

\noindent\textbf{Multi-scaled searching radius.}
PMP-Net searches the point moving path in a coarse-to-fine manner. For each step, PMP-Net reduces the maximum stride to move a point by the power of 10, which is, for step $k$, the displacement $\Delta\bs{p}^k_i$ calculated in Eq.(\ref{eq:displacement}) is limited to $10^{-k+1}\Delta\bs{p}^k_i$. This allows the network converges more quickly during training. And also, the reduced searching range will guarantee the network at next step not to overturn its decision made in the previous step, especially for the long range movements. Therefore, it can prevent the network from making redundant decision during path searching process.

\subsection{Training Loss}
The deformed shape is regularized by the complete ground truth point cloud through Chamfer distance (CD) and Earth Mover Distance (EMD). Following the same notations in Eq.(\ref{eq:emd}), the Chamfer distance is defined as:
\begin{equation}\label{eq:cd}\small
  \mathcal{L}_{CD}({X},\hat{X})=\sum_{\bs{x}\in X}\min_{\hat{\bs{x}}\in\hat{X}}\|\bs{x}-\hat{\bs{x}} \| + \sum_{\hat{\bs{x}}\in \hat{X}}\min_{\bs{x}\in X}\|\hat{\bs{x}}-\bs{x} \|.
\end{equation}
The total loss for training is then given as
\begin{equation}\small
  \mathcal{L}=\sum_{k}\mathcal{L}_{\mathrm{CD}}(P^k,P')+\mathcal{L}_{\mathrm{PMD}},
\end{equation}
where $P^k$ and $P'$ denote the point cloud output by step $k$ and the target complete point cloud, respectively. Note that finding the optimal $\phi$ is extremely computational expensive. In experiments, we follow the simplified algorithm in \cite{yuan2018pcn} to estimate an approximation of $\phi$.

\begin{figure*}[!t]
  \centering
  \includegraphics[width=0.95\textwidth]{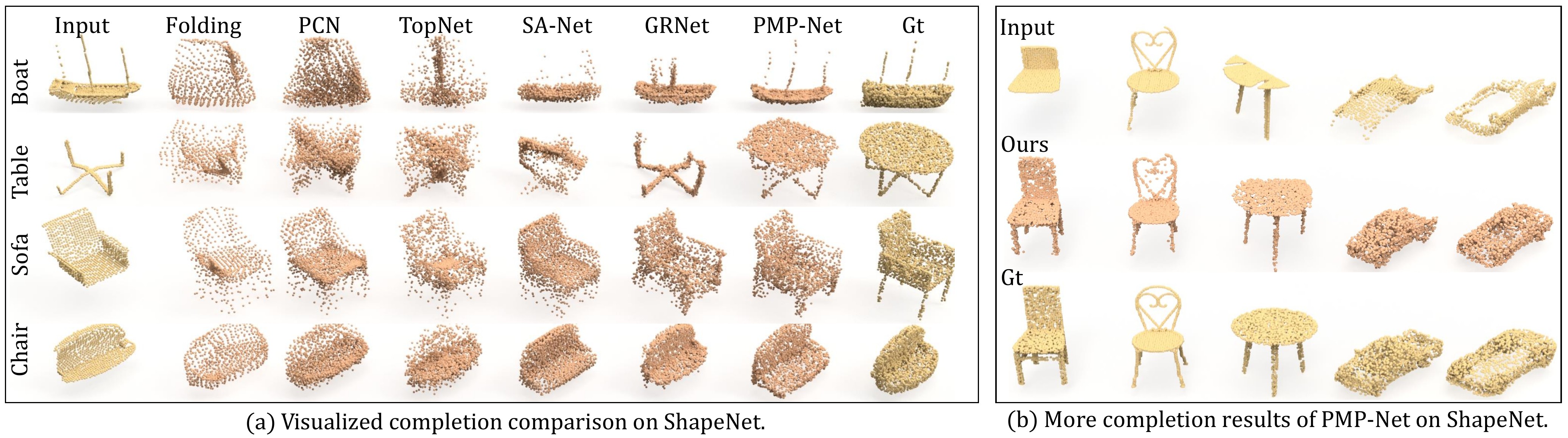}\vspace{-0.2cm}
  \caption{Visualization of point cloud completion results on Completion3D dataset. PMP-Net is compare with other methods in (a), and (b) shows more completion results of PMP-Net.
  }
  \label{fig:shapenet_vis}
\end{figure*}

\section{Experiments}
\subsection{Evaluation on Completion3D Dataset}
\noindent\textbf{Dataset.} We evaluate our PMP-Net on the widely used benchmark of 3D point cloud completion, i.e. \emph{Completion3D} \cite{tchapmi2019topnet}, which is a large-scaled 3D object dataset derived from the ShapeNet dataset. The partial 3D shapes are generated by back-projecting 2.5D depth images from partial views into 3D space.%Completion3D dataset concerns the completion task of sparse point cloud completion, where it generates only one partial view for each complete 3D object in ShapeNet dataset, and samples 2,048 points from the mesh surface for both the complete and partial shapes.
We follow the settings of training/validation/test splits in Completion3D for fair comparison with the other methods.

\noindent\textbf{Evaluation metric.} Following previous studies \cite{tchapmi2019topnet,xie2020grnet,wen2020sa,yuan2018pcn}, we use the per-point L1/L2 Chamfer distance (CD) as the evaluation metric. L1 version is the CD in Eq.(\ref{eq:cd}) averaged by the point number, and L2 version is the CD replacing L1-norm in Eq.(\ref{eq:cd}) by L2-norm.

\noindent\textbf{Quantitative comparison.} The comparison results\footnote{Results are cited from \url{https://completion3d.stanford.edu/results}} of PMP-Net with the other state-of-the-art point cloud completion methods is shown in Table \ref{table:completion3D}, in which the PMP-Net achieves the best performance in terms of average chamfer distance across all categories. Compared with the second best completion method GRNet (in terms of average chamfer distance), PMP-Net achieves better results in 6 out of 8 categories, which proves the better generalization ability of PMP-Net across different shape categories.
%Note that GRNet(16k) is the original results reported in the paper \cite{xie2020grnet}. It was obtained by predicting 16,384 points first and down-sample to 2,048 points, in order to improve point uniformity and reduce the chamfer distance.
%For fair comparison, we also reproduce the results of GRNet(2k) to directly generate 2,048 points and evaluate on Completion3D test set.
As we discussed in Sec.2, GRNet \cite{xie2020grnet} is a voxel aided shape completion method, where the conversion between point clouds and 3D voxel along with applying 3D CNN on voxel data is a time consuming process. Other methods like SA-Net \cite{wen2020sa} in Table \ref{table:completion3D} are typical generative completion methods which are fully based on point clouds, and the nontrivial improvement of PMP-Net over these methods proves the effectiveness of deformation based solution in point cloud completion task.
%Moreover, as one of the latest studies, SA-Net  considers detailed shape prediction

\begin{table*}[!t]\small
\centering
\caption{Point cloud completion on PCN dataset in terms of per-point L1 Chamfer distance $\times 10^{3}$ (lower is better).}
%\resizebox{\linewidth}{!}{
\begin{tabular}{l|c|cccccccc}
\toprule
Methods &Average  &Plane    &Cabinet  &Car   &Chair   &Lamp   &Couch    &Table    &Watercraft      \\ \midrule
FoldingNet  \cite{yang2018foldingnet}   &14.31  &9.49    &15.80    &12.61    &15.55   &16.41    &15.97    &13.65    &14.99   \\
TopNet  \cite{tchapmi2019topnet}     &12.15   &7.61   &13.31   &10.90    &13.82    &14.44      &14.78   &11.22  &11.12   \\
AtlasNet \cite{groueix2018atlasnet}   &10.85   &6.37    &11.94    &10.10    &12.06    &12.37    &12.99    &10.33    &10.61   \\
PCN  \cite{yuan2018pcn}   &9.64 &5.50    &22.70    &10.63    &8.70    &11.00    &11.34    &11.68    &8.59   \\
GRNet \cite{xie2020grnet}  &8.83   &6.45   &10.37   &9.45    &\textbf{9.41}    &7.96      &\textbf{10.51}   &8.44  &8.04   \\
CDN. \cite{wang2020cascaded}  &\textbf{8.51}   &\textbf{4.79}   &\textbf{9.97}   &\textbf{8.31}    &9.49    &8.94      &10.69   &\textbf{7.81}  &8.05   \\
\midrule
PMP-Net(Ours) &8.66  &5.50    &11.10    &9.62    &9.47    &\textbf{6.89}    &10.74    &8.77   &\textbf{7.19} \\
\bottomrule
\end{tabular}
%}
\label{table:PCN}
\end{table*}

\noindent\textbf{Qualitative comparison.} In Figure \ref{fig:shapenet_vis}(a), we visually compare PMP-Net with the other completion methods on Completion3D dataset, from which we can find that PMP-Net predicts much more accurate complete shapes on various input categories, while the other methods may output some failure cases for certain input shapes. For example, the input \emph{table} in Figure \ref{fig:shapenet_vis}(a) loses the entire desktop, and methods like GRNet and SA-Net almost cannot predict the whole desktop, while other methods like FoldingNet \cite{yang2018foldingnet}, PCN \cite{yuan2018pcn} and TopNet \cite{tchapmi2019topnet} intend to repair the desktop but fail to predict a good shape of it. Moreover, the advantage of deformation based PMP-Net over the generative methods can be well proved by the case of \emph{boat} in Figure \ref{fig:shapenet_vis}(a). Generative methods, especially like PCN and TopNet, successfully learn the overall structure of the input boat, but fail to reconstruct even the residual part of it. On the other hand, the deformation based PMP-Net can directly preserve the input shape by moving small amount of point to perform completion on certain areas, and keep the input shape unchanged. In Figure \ref{fig:shapenet_vis}(b), we visualize some more completion results of PMP-Net.

\begin{figure}[!t]
  \centering
  \includegraphics[width=\columnwidth]{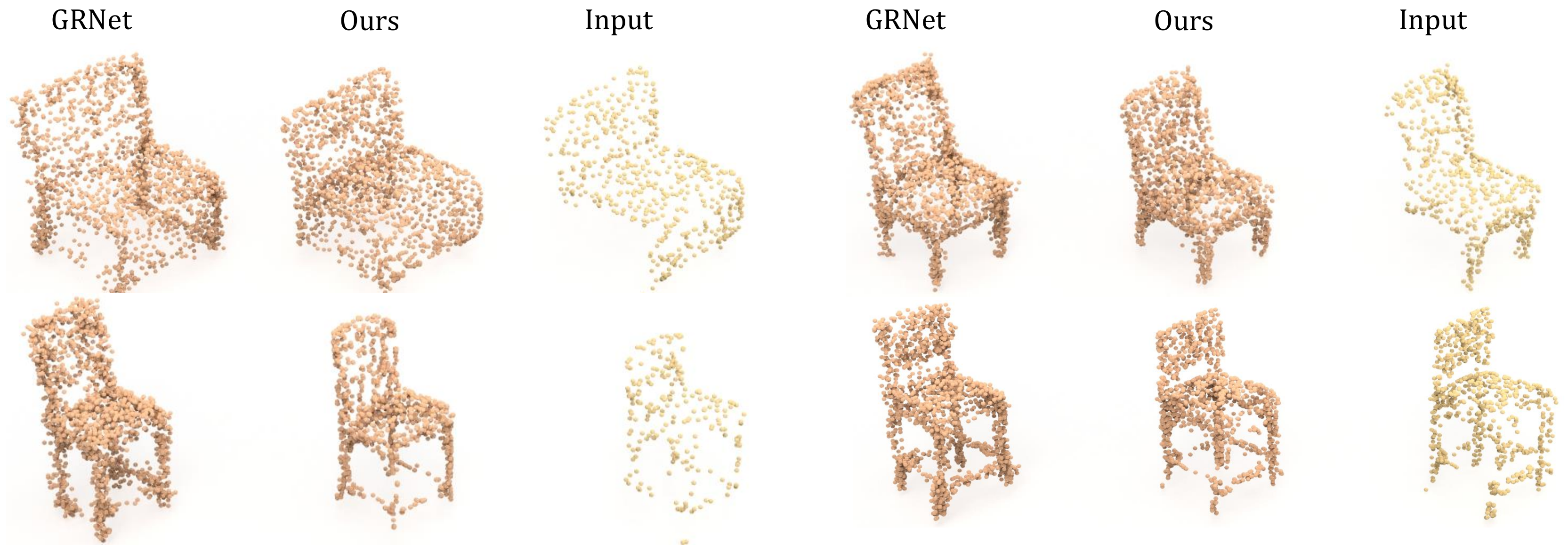}\vspace{-0.4cm}
  \caption{Visual comparison of PMP-Net and GRNet on ScanNet chairs.
  }
  \label{fig:scannet_vis}
\end{figure}

\noindent\textbf{Qualitative comparison on ScanNet chairs.}
To evaluate the generalization ability of PMP-Net on point cloud completion task, we pre-train PMP-Net on Completion3D dataset and evaluate its performance on the chair instances in ScanNet dataset without finetune, and compare with GRNet (which is the second best method in Table \ref{table:completion3D}, and also pre-trained on Completion3D). The visual comparison is shown in Figure \ref{fig:scannet_vis}. The PMP-Net completes shapes with less noise than GRNet, which benefits from its point moving practice. Because the differences in data distribution between Completion3D and ScanNet will inevitably confuse the network, and the point moving based PMP-Net can simply choose to leave those points in residual part of an object to stay at their own place to preserve a better shape, in contrast, generation based GRNet has to predict new points for both residual and missing part of an object.

\subsection{Applying to Dense Point Cloud Completion}
We show that PMP-Net learned on sparse point cloud can be directly applied to the dense point cloud completion. Specifically, we keep training PMP-Net on sparse shape with 2,048 points, and reveal its generalization ability by predicting dense complete shape with 16,384 points on PCN dataset \cite{yuan2018pcn}. PCN dataset is derived from ShapeNet dataset, in which each complete shape contains 16,384 points. The partial shapes have various point numbers, so we first down-sample shapes with more than 2,048 points to 2,048, and up-sample shapes with less than 2048 points to 2048 by randomly copying points.
Since PMP-Net learns to move points instead of generating points, it requires the same number of points in incomplete point cloud and complete one. In order to predict complete shape of 16,384 points, we repeat 8 times of prediction for each shape during testing, with each time moving 2,048 points. Note that PMP-Net is still trained on sparse point clouds with 2,048 points, which are sampled from the dense point clouds of PCN dataset.

The comparison in Table \ref{table:PCN} shows that PMP-Net yields a comparable performance to the state-of-the-art method \cite{wang2020cascaded}, and ranks second on PCN dataset. The result of Wang et al.\cite{wang2020cascaded} is cited from its original paper, while the results of other compared methods are all cited from \cite{xie2020grnet}. Note that most generation based methods (like Wang et al. \cite{wang2020cascaded} and GRNet \cite{xie2020grnet} in Table \ref{table:PCN}) specially designed a coarse-to-fine generation process in order to obtain better performance on dense point cloud completion. In contrast, our PMP-Net trained on 2,048 points can directly generate arbitrary number of dense points by simply repeating the point moving process, and still achieves comparable results to the counterpart methods.

\subsection{Model Analysis}
In this subsection, we analyze the influence of different parts in the PMP-Net. By default, we use the same network sittings in all experiments except for the analyzed part. All studies are typically conducted on the validation set of Completion3D dataset under four categories (i.e. plane, car, chair and table) for convenience.

\noindent\textbf{Analysis of RPA module and PMP loss.} We analyze the effectiveness of RPA module by replacing it with other units in PMP-Net. And for PMP loss, we analyze its effectiveness by removing PMP loss from the network. Specifically, we develop six different variations for comparison: (1) \emph{NoPath} is the variation that removes the RPA module from the network; (2) \emph{Add} is the variation that replaces RPA module with element-wise add layer in the network; (3) RNN, (4) LSTM and (5) GRU are variations that replace RPA module with different recurrent unit; (6) \emph{w/o PMP} is the variation that removes PMP loss from the PMP-Net, where only Chamfer distance is used for training.

The shape completion results are shown in Table \ref{table:analysis_rpa}, from which we can find that baseline RPA module achieves the best performance. The worst results yielded by \emph{Add} variation indicate that directly utilizing history information (paths in previous step) without processing will in return degenerate the network performance, compared to \emph{NoPath} variation. The comparison between \emph{RPA} baseline and \emph{GRU} variation proves the effectiveness of our designation of RPA module, which can give more consideration to the information from current step than GRU unit, and help the network to make more precise decision for point moving.

By comparing baseline variation with \emph{w/o PMP} variation, we can find that PMP loss significantly improves the performance of our network, which is in accordance with our opinion that point moving path should be regularized to better capture the detailed topology and structure of 3D shapes.
\begin{table}[!h]\small
\centering
\caption{Analysis of RPA and PMP loss (baseline marked by ``*'').}
\begin{tabular}{lccccc}
\toprule
Unit. &avg.    &plane   &chair  &car  &table  \\ \midrule
NoPath &11.95 &3.55 &8.30  &16.15 &19.79\\
Add  &12.23  &\textbf{3.32}  &8.10  &16.47 &21.05\\
RNN  &12.12  &3.55  &8.14  &16.19 &20.58\\
LSTM  &11.99  &3.79  &8.09  &15.37 &20.72\\
GRU  &11.87  &3.44  &7.85  &15.44 &20.72\\
w/o PMP &13.66 &4.26 &8.21 &16.69 &25.19 \\
\midrule
baseline* &\textbf{11.58}  &3.42  &\textbf{7.87}  &\textbf{15.88} &\textbf{19.15}\\
\bottomrule
\end{tabular}
\label{table:analysis_rpa}
\end{table}

\noindent\textbf{Effect of multi-step path searching.} In Table \ref{table:step_analysis}, we analyze the effect of different steps for point cloud deformation. Specifically, we fix the ratio of searching radius between each step to $10$, and evaluate the performance of PMP-Net under different step sittings. For example, the searching radius for step=4 is set to $\{1.0, 10^{-1}, 10^{-2}, 10^{-3}\}$, and the searching radius for step=2 is set to $\{1.0, 10^{-1}\}$, respectively. From Table \ref{table:step_analysis}, we can find that deforming point cloud by multiple steps effectively improves the completion performance, by comparing the results of step 1, 2 and 3. On the other hand, the comparison between step 3 and step 4 shows that the performance of multi-step path searching will reach its limitation, because too many steps may cause information redundancy in path searching.
\begin{table}[!h]\small
\centering
\caption{The effect of different steps (baseline marked by ``*'').}
\begin{tabular}{lccccc}
\toprule
Steps. &avg.    &plane   &chair  &car  &table  \\ \midrule
1  &12.26  &3.71  &8.27  &15.59 &21.48\\
2  &11.90  &3.47  &7.95  &15.66 &20.53\\
3*  &\textbf{11.58}  &3.42  &\textbf{7.87}  &\textbf{15.88} &\textbf{19.15}\\
4  &11.67  &\textbf{3.39}  &7.91  &15.89 &19.48\\
\bottomrule
\end{tabular}
\label{table:step_analysis}
\end{table}

%\subsubsection{Analysis of Searching Radius}
\noindent\textbf{Analysis of searching radius.} By default, we decrease the searching radius for each step by the ratio of 10. In Table \ref{table:radius_analysis}, we analyze different sittings of searching radius and evaluate their influence to the performance of PMP-Net. We additionally test two different strategies to perform point moving path searching, i.e. the strategy without decreasing searching radius ($[1.0,1.0,1.0]$ for each step), and the strategy with smaller decreasing ratio ($[1.0,0.5,0.25]$). The baseline result is the default sitting of PMP-Net ($[1.0,0.1,0.01]$ for each step).  Table \ref{table:radius_analysis} shows that PMP-Net achieves the worst performance at $[1.0,1.0,1.0]$, which proves the effectiveness of the strategy to decrease searching radius. And when comparing strategy of $[1.0,0.5,0.25]$ with $[1.0,0.1,0.01]$, we can find that decreasing searching radius with larger ratio can improve the model performance, because larger ratio can better prevent the network from overturning the decisions in previous steps. We also note that when the decreasing ratio goes to large, the PMP-Net will approximate the behavior of network with step=1 in Table \ref{table:step_analysis}, which can in return harm the performance of shape completion.
\begin{table}[!h]\small
\centering
\caption{The effect of searching radius (baseline marked by ``*'').}
\resizebox{\columnwidth}{!}{\begin{tabular}{lccccc}
\toprule
Radius. &Avg.    &Plane   &Chair  &Car  &Table  \\ \midrule
$[1.0,1.0,1.0]$  &12.01  &3.61  &8.22  &16.44 &19.79\\
$[1.0,0.5,0.25]$  &11.77  &\textbf{3.36}  &8.01  &15.92 &19.79\\
$[1.0,0.1,0.01]$*  &\textbf{11.58}  &3.42  &\textbf{7.87}  &\textbf{15.88} &\textbf{19.15}\\
\bottomrule
\end{tabular}}
\label{table:radius_analysis}
\end{table}

\begin{figure}[!t]
  \centering
  \includegraphics[width=\columnwidth]{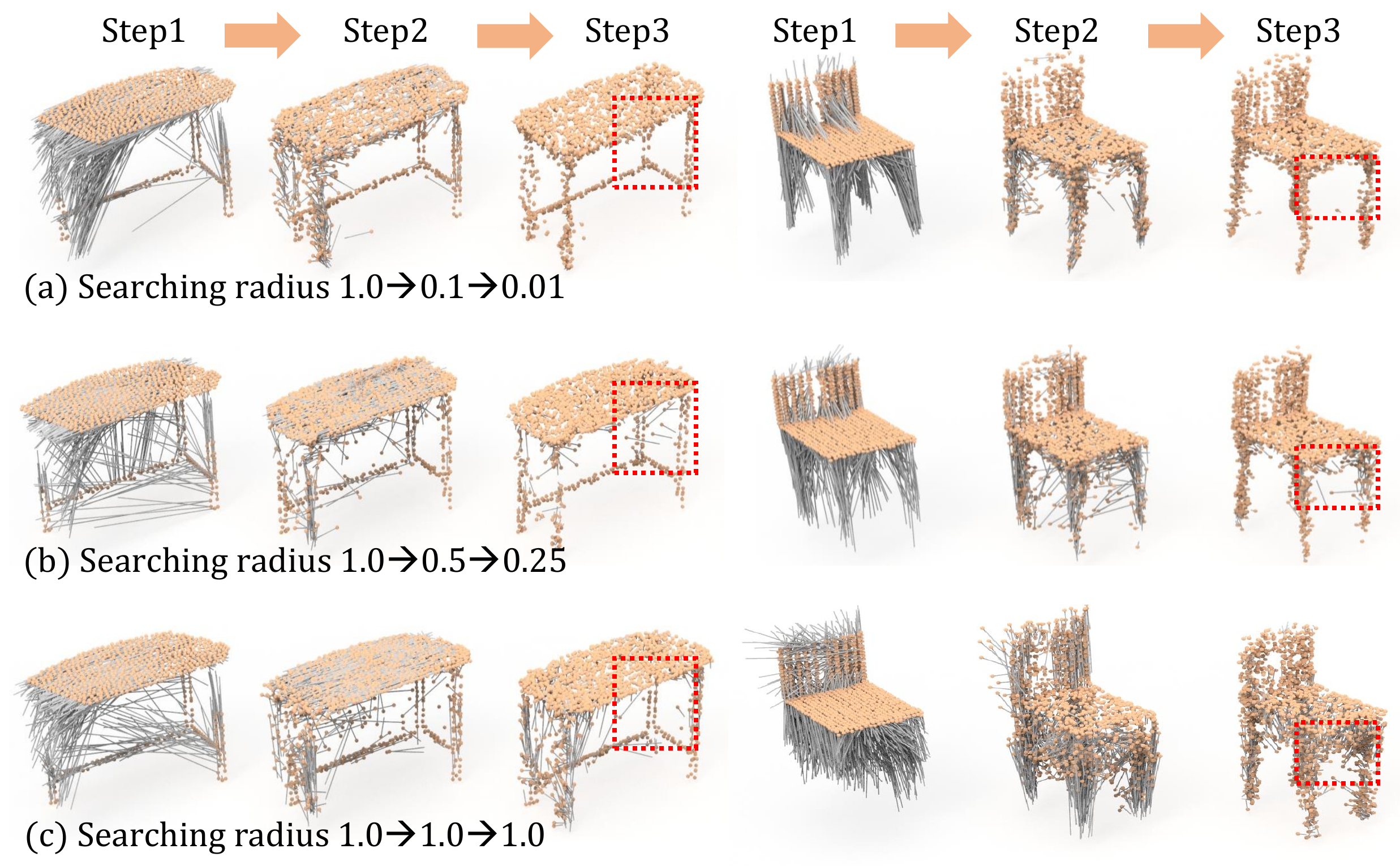}\vspace{-0.4cm}
  \caption{Illustration of deformation process in each step under different strategies of searching radius.
  }
  \label{fig:radius_analysis}
\end{figure}

\noindent\textbf{Visual analysis of point moving path under different radius.}
In Figure \ref{fig:radius_analysis}, we visualize the searching process under different strategies of searching radius in Table \ref{table:step_analysis}. By analyzing the deformation output in step 1, we can find that PMP-Net with a coarse-to-fine searching strategy can learn to predict a better shape at early step, where the output of step 1 in Figure \ref{fig:radius_analysis}(a) is more complete and tidy than the ones in Figure \ref{fig:radius_analysis}(b) and Figure \ref{fig:radius_analysis}(c). Moreover, a better overall shape predicted in the early stage will enable the network focus on refining a better detailed structure of point cloud, which can be concluded from the comparison of step 3 in Figure \ref{fig:radius_analysis}, where the region highlighted by red rectangles in Figure \ref{fig:radius_analysis}(a) is much better than the other two subfigures.

\section{Conclusions}
In this paper, we propose a novel PMP-Net for point cloud completion by multi-step shape deformation. By moving points from source to target point cloud with multiple steps, PMP-Net can consistently refine the detailed structure and topology of the predicted shape, and establish the point-level shape correspondence between the incomplete and the complete shape. In experiments, we show the superioty of PMP-Net by comparing with other methods on the Completion3D benchmark and PCN dataset.
%go deep to quantitatively and visually analyze the effectiveness of our proposed RPA module and path searching strategies.

{\small
\bibliographystyle{ieee_fullname}
\bibliography{ref}
}
\newpage
\appendix
\section{Detailed Settings}
We use the single scale grouping (SSG) version of PointNet++ and its feature propagation module as the basic framework of PMP-Net. The detailed architecture of each part is described in Table \ref{table:encoder} and Table \ref{table:fp_module}, respectively.
\begin{table}[!h]\small
\centering
\caption{The detailed structure of encoder.}
\begin{tabular}{lcccc}
\toprule
Level &\#Points    &Radius   &\#Sample  &MLPs  \\ \midrule
1  &512  &0.2  &32  &$[64,64,128]$\\
2  &128  &0.4  &32  &$[128,128,256]$ \\
3  &-  &-  &- &$[256,512,1024]$ \\
\bottomrule
\end{tabular}
\label{table:encoder}
\end{table}

In Table \ref{table:encoder}, ``\#Points'' denotes the number of down-sampled points, ``Radius'' denotes the radius of ball query, ``\#Sample'' denotes the number of neighbors points sampled for each center point, ``MLPs'' denotes the number of output channels for MLPs in each level of encoder.

\begin{table}[!h]\small
\centering
\caption{The detailed architecture of feature propagation module.}
\begin{tabular}{lccc}
\toprule
Level &1 &2 &3  \\ \midrule
MLPs  &$[256,256]$ &$[256,128]$ &$[128,128,128]$\\
\bottomrule
\end{tabular}
\label{table:fp_module}
\end{table}

\noindent\textbf{Training details.} We use AdamOptimizer to train PMP-Net with an initial learning rate $10^{-3}$, and exponentially decay it by 0.5 for every 20 epochs. The training process is accomplished using a single NVDIA GTX 2080TI GPU with a batch size of 24. PMP-Net takes 150 epochs to converge on both PCN and Completion3D dataset. We scale all input training shapes of Completion3D by 0.9 to avoid points out of the range of \emph{tanh} activation.

\section{More Experiments}
\subsection{Dimension of Noise Vector.} The noise vector in Eq.(1) in our paper is used to push the points to leave their original place. In this section, we analyze the dimension and the standard deviation of the noise, which may potentially decide the influence of the noise to the points. Because either the dimension or the standard deviation of the noise vector decreases to 0, there will be no disturbance in the network. On the other hand, larger vector dimension or standard deviation will cause larger disturbance in the network. In Table \ref{table:analysis_noiseDim}, we first analyze the influence of dimension of noise vector. By comparing 0-dimension result with others, we can draw conclusion that the disturbance caused by noise vector is important to learn the point deformation. And by analyzing the performance of different length of noise vector, we can find that the influence of vector length is relatively small, compared with the existence of noise vector.
%In Table \ref{table:analysis_noiseStd}, we further analyze the standard deviation of noise vector.

\begin{table}[!h]\small
\centering
\caption{The effect of noise dimension (baseline marked by ``*'').}
\begin{tabular}{lccccc}
\toprule
Dim. &avg.    &plane   &chair  &car  &table  \\ \midrule
0  &14.56  &4.39  &10.48  &19.01 &24.33\\
8  &11.85  &3.28  &7.95  &15.65 &20.50\\
16  &11.68  &3.44  &\textbf{7.86}  &\textbf{15.22} &20.19\\
32*  &\textbf{11.58}  &3.42  &7.87  &15.88 &\textbf{19.15}\\
64  &\textbf{11.58}  &\textbf{3.14}  &7.96  &16.01 &19.17\\
\bottomrule
\end{tabular}
\label{table:analysis_noiseDim}
\end{table}

\subsection{Standard Deviation of Noise Distribution.} In Table \ref{table:analysis_noiseStd}, we show the completion results of PMP-Net under different standard deviations of noise vector. Similar to the analysis of vector dimension, we can draw conclusion that larger disturbance caused by bigger standard deviation will help the network achieve better completion performance. The influence of noise vector becomes weak when the standard deviation reaches certain threshold (around $10^{-1}$ according to Table \ref{table:analysis_noiseStd}).
\begin{table}[!h]\small
\centering
\caption{The effect of standard deviation (baseline marked by ``*'').}
\begin{tabular}{lccccc}
\toprule
Stddev.     &Avg.    &Plane   &Chair  &Car  &Table  \\ \midrule
$10^{-2}$   &11.89  &3.32  &8.15  &16.42 &19.58\\
$10^{-1}$   &\textbf{11.56}  &3.58  &7.78  &15.47 &19.41\\
$1.0$*       &11.58  &3.42  &\textbf{7.87}  &15.88 &\textbf{19.15}\\
$10$        &11.62  &\textbf{3.35}  &7.88  &\textbf{15.29} &19.95\\
\bottomrule
\end{tabular}
\label{table:analysis_noiseStd}
\end{table}

\subsection{Visual Analysis of Multi-step Searching.}
\begin{figure}[!t]
  \centering
  \includegraphics[width=\columnwidth]{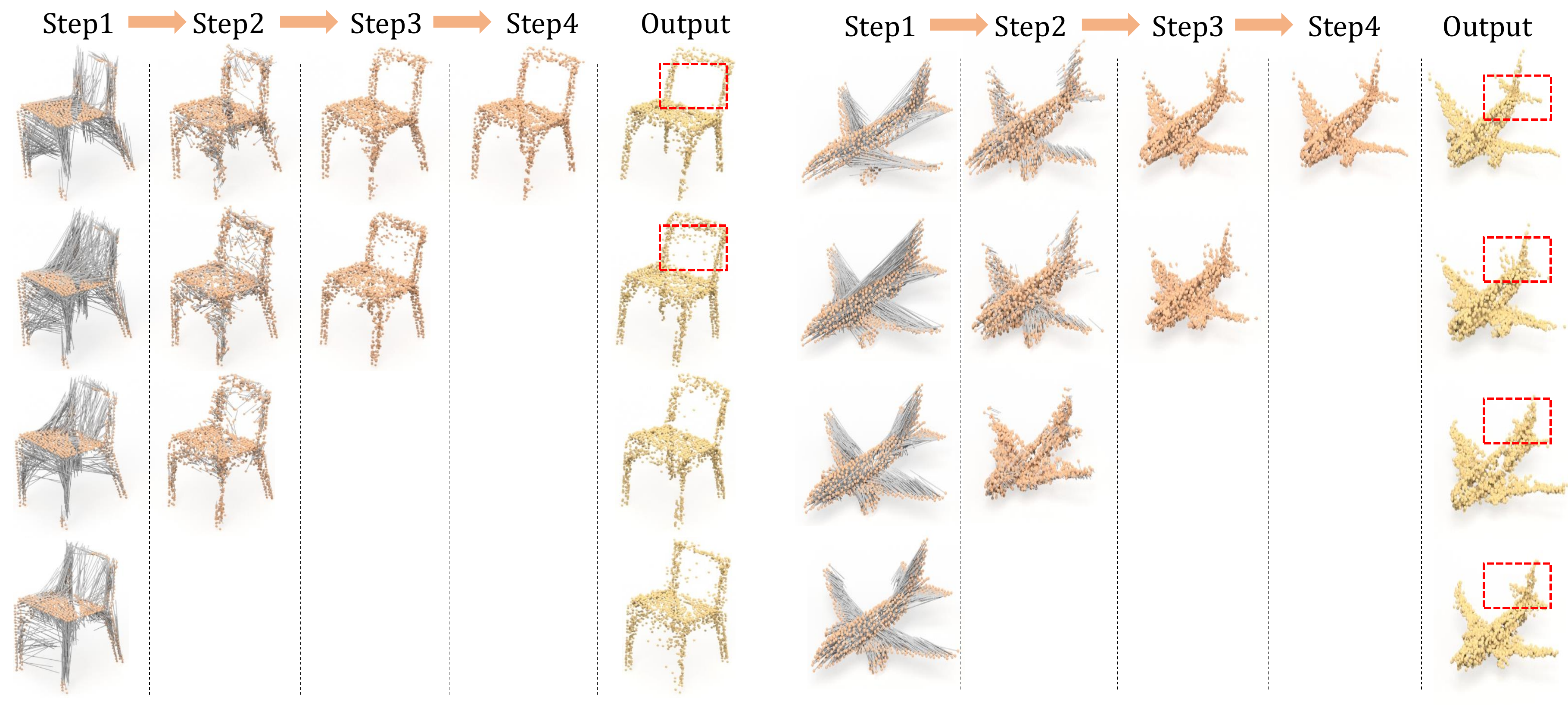}\vspace{-0.4cm}
  \caption{Illustration of multi-step searching under different searching steps. The first row is 4-step completion, and the second row is 3-step completion, and so on.
  }
  \label{fig:step_analysis}
\end{figure}

We visualize the point deformation process under different searching step sittings in Figure \ref{fig:step_analysis}. Comparing the 4-step searching in the top-row with the other three sittings, the empty space on the chair back is shaped cleaner as highlighted by rectangles, which proves the effectiveness of multi-step searching to consistently refine the shape.

\subsection{Visualization of Completion Results on PCN dataset.}
\begin{figure*}[!t]
  \centering
  \includegraphics[width=0.95\textwidth]{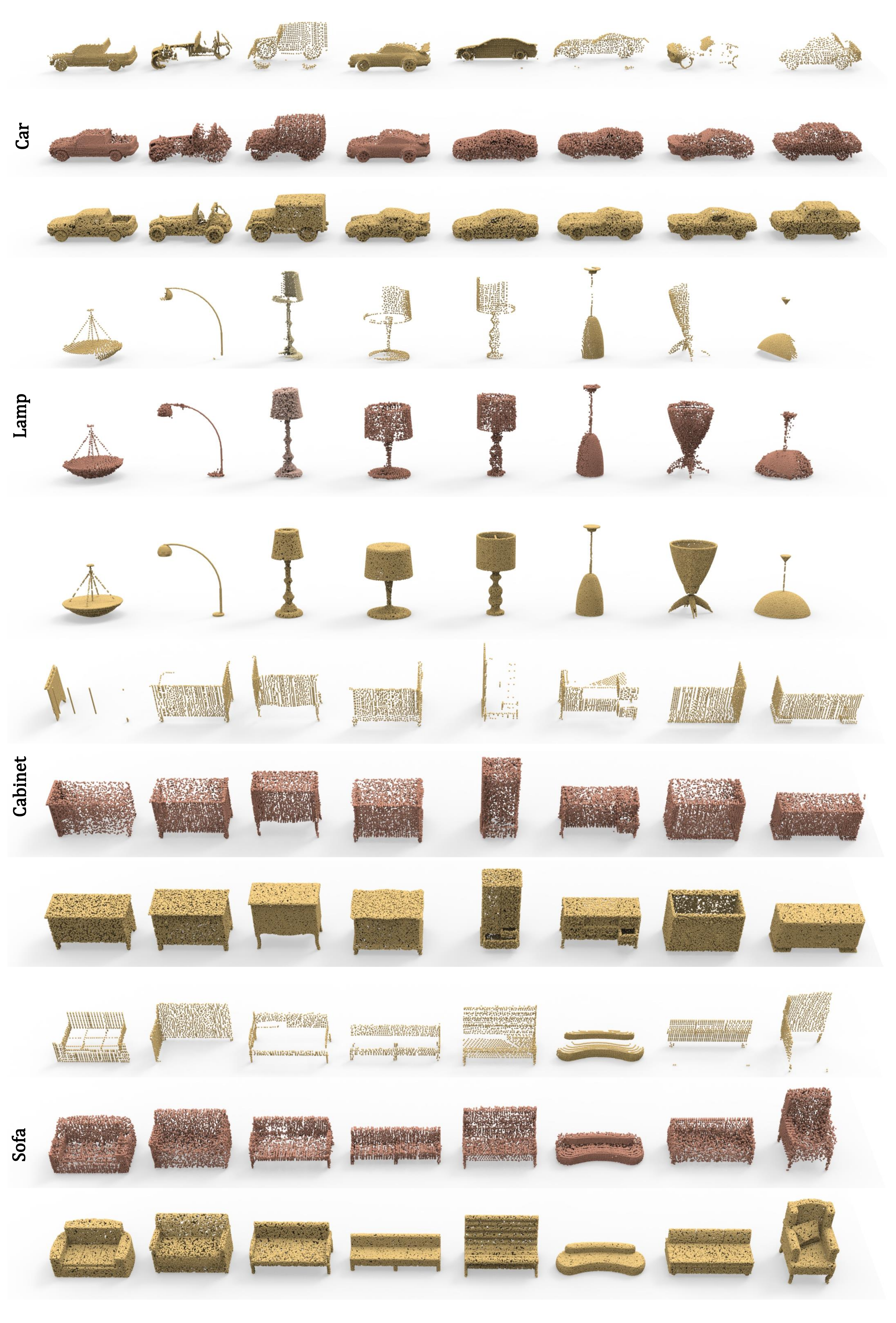}
  \caption{Illustration of shape completion on PCN dataset. For each category, the first row is the input incomplete shape, the second row is the predicted complete shape and the third row is the ground truth.
  }
  \label{fig:pcn1}
\end{figure*}

\begin{figure*}[!t]
  \centering
  \includegraphics[width=0.95\textwidth]{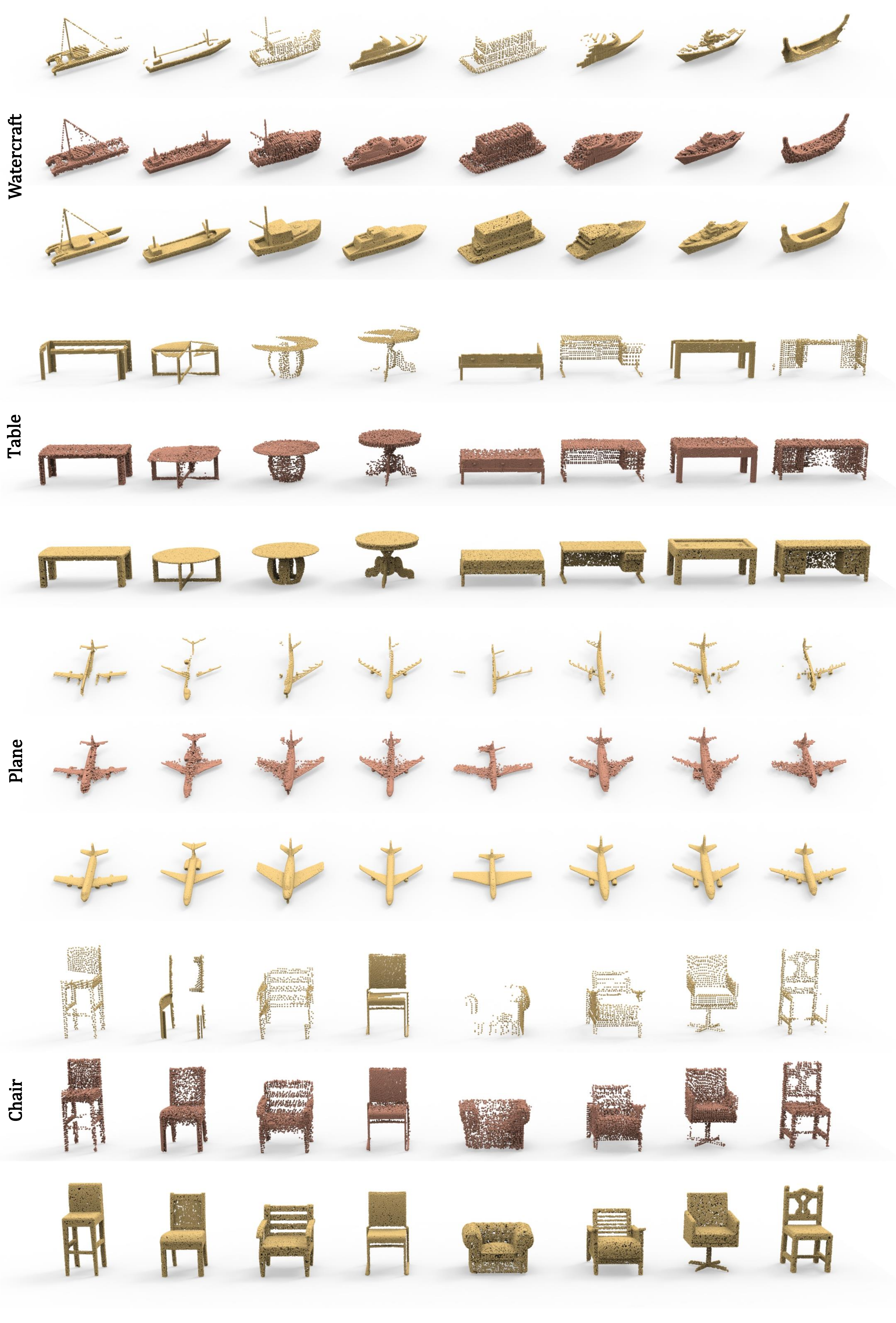}
  \caption{Illustration of shape completion on PCN dataset. For each category, the first row is the input incomplete shape, the second row is the predicted complete shape and the third row is the ground truth.
  }
  \label{fig:pcn2}
\end{figure*}

In Figure \ref{fig:pcn1} and Figure \ref{fig:pcn2}, we supplement results of shape completion on PCN dataset under each categories. For each category, the first row is the input incomplete shape, the second row is the predicted complete shape and the third row is the ground truth.

\end{document}